\documentclass{article}

\usepackage[preprint]{neurips_2026}

\usepackage[utf8]{inputenc} 
\usepackage[T1]{fontenc}    
\usepackage{hyperref}       
\usepackage{url}            
\usepackage{booktabs}       
\usepackage{amsfonts}       
\usepackage{nicefrac}       
\usepackage{microtype}      
\usepackage[table]{xcolor}         
\usepackage{graphicx}
\usepackage{wrapfig}
\usepackage{adjustbox}

\usepackage{algorithm}
\usepackage{algpseudocode}
\usepackage{wrapfig}
\usepackage{caption}
\usepackage{amsmath}
\usepackage{cleveref}
\usepackage{multirow}
\usepackage{enumitem}

\algrenewcommand\algorithmiccomment[1]{$\triangleright$ {\color{gray}#1}}
\crefname{appendix}{Appendix}{Appendices}
\Crefname{appendix}{Appendix}{Appendices}

\title{
Can Model Merging Improve Aggregation in DiLoCo?
}

\author{%
    \parbox{\textwidth}{\centering
      \vspace{0.75cm}
      Stefan Horoi$^{1,2}$\thanks{Correspondence to: \texttt{stefan.horoi@umontreal.ca}, \texttt{eugene.belilovsky@concordia.ca}\\
      \hspace*{1.8em}Code available at: \url{https://github.com/shoroi/isoloco}
      }
      \qquad Benjamin Thérien$^{1,2}$
      \qquad Guy Wolf$^{1,2}$
      \qquad Eugene Belilovsky$^{2,3}$
      \vspace{5pt}\\
      \textnormal{$^1$Université de Montréal \quad $^2$Mila -- Quebec AI Institute \quad $^3$Concordia
  University}
    }
  }

\begin{document}

\maketitle

\begin{abstract}
Model merging techniques, which aggregate independently finetuned models into one to combine their capabilities, have become a topic of significant interest in recent years, with a broad array of methods having been proposed to tackle this problem. Simultaneously, an emerging trend in distributed learning has been the use of methods such as local SGD and DiLoCo, which greatly reduce communication costs by periodically aggregating the independently trained local models. However, these communication-efficient methods have been shown to degrade in performance relative to the FLOP-matched data-parallel gold standard as the number of independent local models grows and as the number of local training steps before global communication is increased. In this work, we draw an explicit analogy between the pseudo-gradient aggregation step in local SGD/DiLoCo and task arithmetic-based model merging, establishing a straightforward way to utilize merging methods in the context of distributed optimization. We then evaluate multiple state-of-the-art model merging methods in this setting and identify one method in particular, Iso-C, as a promising approach for improving DiLoCo. We find that DiLoCo SGD with Iso-C aggregation outperforms not only simple pseudo-gradient averaging but even the momentum-based DiLoCo, despite lacking a momentum mechanism itself. Building on this finding, we propose IsoLoCo, which adapts Iso-C for distributed training by equipping it with Nesterov momentum. Our empirical evaluations on language model pre-training across varying numbers of local workers show that IsoLoCo significantly outperforms DiLoCo, with the gap between them widening as the number of workers increases. This advantage remains present across model sizes and inner step counts, confirming that merging-inspired aggregation is an effective strategy for low-communication distributed training.
\end{abstract}


\section{Introduction}
%
%
The recent success of large language models (LLMs) has relied in large part on the scaling of model parameters, dataset size and amount of compute used for training. However, training models with parameter counts in the billions or trillions on internet-scale datasets comes with significant engineering challenges. The standard practice is to use data-parallel training, where multiple replicas of the model process mini-batches of data in parallel and their gradients are aggregated to update model parameters via gradient-based optimization.
This requires large clusters of tightly co-located accelerators connected with high-bandwidth links, along with careful engineering to orchestrate communication and ensure efficient resource utilization. Furthermore, such synchronized training is vulnerable to device failures and makes poor use of heterogeneous hardware with varying processing speeds or network topologies.

%

{\parfillskip=0pt Recently,~\cite{douillard2024diloco} introduced Distributed Low-Communication (DiLoCo) training, a variant of FedAvg~\cite{pmlr-v54-mcmahan17a-fedavg} and FedOpt~\cite{reddi2021fedopt} with AdamW~\cite{loshchilov2018adamw} as the inner optimizer and stochastic gradient descent (SGD) with Nesterov momentum \cite{Nesterov1983AMF, sutskever2013init_momentum} as the outer optimizer. DiLoCo enables the training of deep learning models on multiple ``islands'' of poorly connected computational clusters.\par}

\begin{wrapfigure}{r}{0.45\linewidth}
  \centering
  \vspace{-12pt}
  \includegraphics[width=\linewidth]{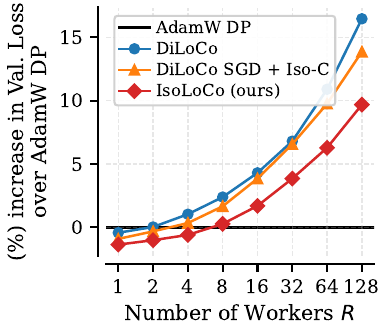} \vspace{-15pt}
  \caption{\textbf{IsoLoCo scales better with workers than DiLoCo.} Final validation loss is shown versus worker count for a 178M LLaMa-2 model in a FLOP-matched setting.
  \vspace{-10pt}
  }
  \label{fig:workerscaling}
\end{wrapfigure}

Each cluster, or \emph{worker}, trains its own model replica through multiple local optimization steps before sharing the accumulated parameter changes, i.e. \emph{pseudo-gradients}, with the remaining clusters for a global optimization step. 
Performing many \emph{inner} optimization steps between each \emph{outer} synchronization allows DiLoCo to reach comparable final performance to data-parallel training while communicating significantly less.
While highly effective at reducing communication costs and even surpassing data-parallel training at small worker counts, DiLoCo’s final performance degrades in FLOP-matched settings as the number of workers or inner steps increases, making it difficult to scale without losing FLOP-efficiency~\cite{charles2026scaling_laws_diloco,therien2026muloco,douillard2024diloco,sarfi2025sparseloco}.
%


In parallel, as strong foundation models have been built and their weights made accessible to the public, these models have been adapted to many downstream tasks, with fine-tuned versions published on public repositories such as HuggingFace~\cite{wolf2019huggingfacetransformers} and AdapterHub~\cite{pfeiffer-etal-2020-adapterhub}. Model merging methods have emerged as a powerful way to leverage such checkpoints, aggregating them into one model with the combined capabilities of its constituents.
Recently, \emph{task arithmetic}~\cite{ilharco2023_task-arithmetic} has emerged as a simple yet powerful paradigm for model merging. Rather than directly aggregating expert model parameters, e.g. by averaging them, \emph{task vectors} are computed to isolate the difference between each expert's fine-tuned parameters and the pre-trained initialization. These task vectors are then aggregated, scaled, and added back to the pre-trained model parameters, often yielding significant performance gains.
However, despite these advances, the merged model often falls short of each fine-tuned expert on its corresponding task, even while achieving better overall multi-task performance than any individual expert.
This merging performance gap is generally attributed to task, or parameter, \emph{interference}~\cite{yadav2023tiesmerging,Gargiulo_2025_TSV-Merging,marczak2025isoc}: a phenomenon in which the accumulated parameter updates induced by training on different tasks can conflict during merging.
%
%
Task interference is known to worsen as the number of models being merged increases~\cite{ilharco2023_task-arithmetic, yadav2023tiesmerging, davari2024breadcrumbs, marczak2025isoc} and with longer fine-tuning duration~\cite{zhou2025on-task-vectors-gradients,horoi2025less-is-more}, motivating a wide range of task-arithmetic-based merging methods aimed at reducing negative interference~\cite{yu2024dare, Gargiulo_2025_TSV-Merging, marczak2025isoc, yang2026model_merging}.

At first glance, low-communication distributed training methods such as DiLoCo and model merging methods appear to address very different problems. DiLoCo is motivated by the need to pre-train large models on massive datasets across multiple, poorly connected compute pools. In this setting, workers optimize the same objective, typically next-token prediction, on disjoint subsets of the same training dataset. By contrast, model merging is usually framed as a post-training technique: multiple checkpoints from a shared base model, fine-tuned on different data distributions and/or objectives, are combined into a single model intended to inherit their capabilities.
Despite these differences, the two families of methods share a common underlying structure. In DiLoCo, the quantities communicated by workers at each outer optimization step are pseudo-gradients: parameter differences between the model at the previous global synchronization point and the locally updated replicas obtained after several inner optimization steps. From the perspective of model merging, these pseudo-gradients can be interpreted as task vectors, where the reference model is the checkpoint from the previous outer step and the ``experts'' are the local replica checkpoints after inner training. The main distinction is that DiLoCo repeats this train-and-aggregate procedure over many rounds, whereas standard model merging typically performs a single, often more substantial, round of specialization followed by one aggregation step.
This connection also suggests that the performance degradation observed in DiLoCo as the number of workers or inner optimization steps increases is directly analogous to the negative task interference studied in the model merging literature. This parallel motivates viewing low-communication distributed training through the lens of model merging, and raises the question of whether merging methods designed to reduce interference can improve DiLoCo-style training.
%

\begin{figure}
    \centering
    \includegraphics[width=1\linewidth]{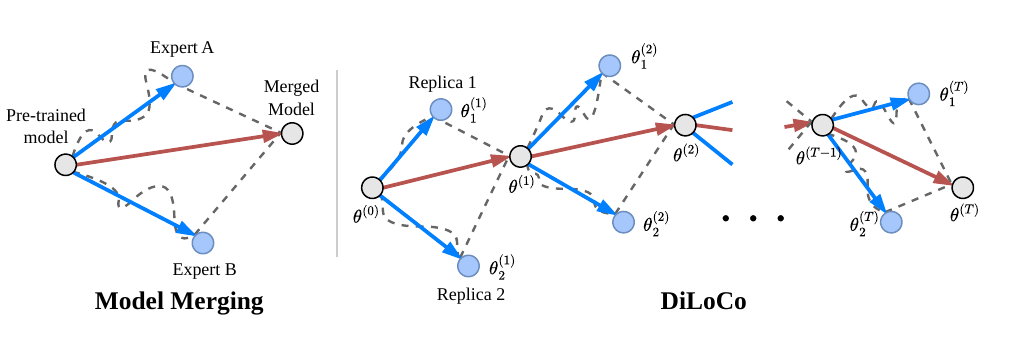}
    \caption{\textbf{DiLoCo as iterative model merging.}
    Dots represent model checkpoints, with blue dots denoting locally trained models: fine-tuned experts in model merging and worker replicas in DiLoCo respectively. Blue arrows are local updates---task vectors in merging and pseudo-gradients in DiLoCo---measuring displacement from the shared initialization. Red arrows denote aggregation: a one-shot merge on the left, and an outer-optimizer update on the right. Formally, with SGD, DiLoCo's outer step reduces to task-arithmetic merging (\Cref{eq:task_arithmetic,eq:diloco_task_arithmetic}), applied iteratively.}
    \vspace{-12pt}
\end{figure}

In this work, we bridge these two closely related but largely separate lines of research. While DiLoCo has recently emerged as an effective approach for distributed pre-training across poorly connected compute pools, its global aggregation step remains simple, typically averaging the pseudo-gradients produced by each worker. In contrast, recent model merging methods propose more sophisticated aggregation mechanisms explicitly designed to reduce negative interference when combining task vectors from multiple independently fine-tuned models, including challenging settings with many experts or long fine-tuning durations.
Building on the structural correspondence identified above, which we make explicit in \Cref{ss:direct_analogy}, we investigate whether model merging techniques can serve as drop-in replacements for DiLoCo's pseudo-gradient aggregation step. Specifically, we ask whether such merging-inspired aggregation can better preserve performance as the number of workers and inner optimization steps increases, two regimes in which low-communication training is known to degrade.
Our contributions are as follows:


\begin{itemize}[leftmargin=2em, itemsep=0pt, topsep=0pt]
    \item We establish a direct analogy between DiLoCo's pseudo-gradient aggregation step and task-arithmetic-based model merging (\Cref{ss:direct_analogy}), showing that model merging methods can be interpreted as alternative aggregation rules for low-communication distributed training. 
    \item Guided by this connection, we systematically benchmark state-of-the-art merging methods as drop-in aggregation mechanisms for DiLoCo-style training in \Cref{s:merging_improves_diloco}, identifying Iso-C~\cite{marczak2025isoc} as offering the strongest performance at reasonable computational cost, even surpassing DiLoCo despite not having a momentum mechanism itself (\Cref{fig:workerscaling} and \Cref{tab:valid-loss-workers}).
    \item We propose \textbf{IsoLoCo} in \Cref{s:isoloco}, a practical adaptation of Iso-C to low-communication distributed training equipped with Nesterov momentum. Through extensive experiments across worker counts (up to $R=128$), number of inner steps, and model sizes, we show that IsoLoCo consistently outperforms DiLoCo and substantially narrows the gap to data-parallel training.
    \item We develop a fast Newton--Schulz-based implementation of IsoLoCo that replaces SVD with iterative orthogonalization and an RMS proxy for the singular-value mean, yielding up to two orders of magnitude faster orthogonalization (${\approx}15\%$ end-to-end speed-up) with negligible loss degradation (${\sim}0.45\%$).
    \item Following recent work that uses Muon as DiLoCo's inner optimizer (MuLoCo)~\cite{therien2026muloco}, we show that IsoLoCo composes with a Muon inner optimizer to further improve worker-count scaling, achieving the lowest validation loss among all methods considered for $R\geq32$, while IsoLoCo with an AdamW inner optimizer matches or surpasses MuLoCo at high worker counts, at a fraction of the orthogonalization cost (Section~\ref{s:muon_inner}).
\end{itemize}


\section{Background and Preliminaries}
\subsection{Distributed Low-Communication Training}\label{s:diloco_background}
The ideas behind DiLoCo can be traced back to federated learning, where training data is distributed across many consumer devices and cannot be centrally aggregated due to privacy or bandwidth constraints.
Federated Averaging~(FedAvg)~\cite{pmlr-v54-mcmahan17a-fedavg} addresses this problem by broadcasting a global model to participating clients, letting each perform several local SGD steps, and aggregating the resulting updates through a weighted average.
When each client takes exactly one step with the same batch size, FedAvg reduces to synchronous data-parallel SGD.
Subsequent work has addressed FedAvg's shortcomings, such as difficulty of hyperparameter tuning and convergence issues, by incorporating adaptive server-side optimizers such as Adam~\citep{kingma2015adammethodstochasticoptimization} in the FedOpt framework~\citep{reddi2021fedopt}.

\begin{wrapfigure}{r}{0.5\textwidth}
\vspace{-1em}
\centering
\begin{minipage}{0.50\textwidth}
\hrule height 0.8pt
\vspace{0.5ex}
\captionof{algorithm}{DiLoCo}
\label{alg:diloco}
\vspace{-5pt}
\hrule height 0.4pt
\begin{algorithmic}[1]
\Require Initial model $\theta^{(0)}$
\Require $R$ workers, Data shards $\{\mathcal{D}_1, \ldots, \mathcal{D}_R\}$
\Require Optimizers $\mathrm{InnerOpt}$ and $\mathrm{OuterOpt}$
\For{outer step $t = 1, \ldots, T$}
    \Statex \Comment{Inner optimization}
    \For{worker $i = 1, \ldots, R$ \textbf{in parallel}}
        \State $\theta_i^{(t)} \gets \theta^{(t-1)}$
        \For{inner step $h = 1, \ldots, H$}
            \State $x \sim \mathcal{D}_i$
            \State $\mathcal{L} \gets f(x, \theta_i^{(t)})$
            \State $\theta_i^{(t)} \gets
            \mathrm{InnerOpt}\bigl(\theta_i^{(t)}, \nabla \mathcal{L}\bigr)$
        \EndFor
    \EndFor
    \Statex \Comment{Average pseudo-gradients}
    \State $\Delta^{(t)} \gets \frac{1}{R} \sum_{i=1}^{R}
    \bigl(\theta^{(t-1)} - \theta_i^{(t)}\bigr)$
    \Statex \Comment{Outer optimization}
    \State $\theta^{(t)} \gets
    \mathrm{OuterOpt}\bigl(\theta^{(t-1)}, \Delta^{(t)}\bigr)$
\EndFor
\end{algorithmic}
\vspace{0.75ex}
\hrule height 0.8pt
\end{minipage}
\vspace{-1em}
\end{wrapfigure}

DiLoCo~\cite{douillard2024diloco} adapts this idea to large-scale language-model pretraining across $R$ workers, or ``islands'' of compute. As shown in \Cref{alg:diloco}, each worker receives the current global model and independently trains a local replica for $H$ inner steps using an inner optimizer ($\mathrm{InnerOpt}$, often AdamW~\cite{loshchilov2018adamw}). After the inner loop, each worker computes a pseudo-gradient capturing its local parameter changes, and these are averaged across workers. The global model is then updated with an outer optimizer ($\mathrm{OuterOpt}$, typically SGD with Nesterov momentum~\citep{Nesterov1983AMF, sutskever2013init_momentum}), applied to this averaged pseudo-gradient. Unlike standard FedOpt, DiLoCo retains inner optimizer states across outer rounds rather than resetting them. Since workers communicate only once every $H$ steps, DiLoCo reduces global synchronization by roughly a factor of $H$ relative to data-parallel, making it attractive for training across weakly connected or geographically distributed clusters.

Subsequent work has established that DiLoCo scales predictably with model size and can match or improve over data-parallel training under fixed compute budgets~\cite{charles2026scaling_laws_diloco}. However, the same studies show that increasing the number of workers $R$ degrades performance~\cite{charles2026scaling_laws_diloco,therien2026muloco}. Similarly, increasing the number of inner steps $H$ leads to worse performance, as local replicas diverge further before aggregation, making the averaged pseudo-gradient less aligned with the update that tightly synchronized training would produce. This limitation is central to our work: DiLoCo's communication savings come at the cost of aggregation quality.

\subsection{Model Merging and Task Arithmetic}
Model merging is a means of combining multiple model checkpoints, fine-tuned from the same pre-trained model, into one with the goal of combining the constituent models' capabilities. It is often justified through the lens of \emph{linear mode connectivity}~\cite{garipov2018_fge, draxler2018_no-barriers, pmlr-v119-frankle20a_lmc_lth}: models fine-tuned from a common pre-trained checkpoint tend to lie in the same basin of the loss landscape \cite{neyshabur2020transfer}, meaning their parameters can be averaged without incurring high loss. However, naive parameter averaging often yields poor results in practice.
A seminal improvement is \textit{task arithmetic} (TA)~\cite{ilharco2023_task-arithmetic}, which operates on \emph{task vectors} rather than raw parameters. Let $\theta_0$ denote the pre-trained parameters and $\theta_t$ those of the model fine-tuned on task $t\in\mathcal{T}$. The task vector $\tau_t=\theta_t-\theta_0$ captures the parameter changes induced by fine-tuning, and models are merged as:
\begin{equation}\label{eq:task_arithmetic}
    \operatorname{Merge}(\theta_0, \{\theta_t\}_{t\in\mathcal{T}}) = \theta_0+\lambda\sum\limits_{t\in\mathcal{T}}\tau_t,
\end{equation}
where $\lambda$ is a tunable hyper-parameter. Task-arithmetic based merging significantly improves merged model performance and has thus become the dominant paradigm in this area of research. Nevertheless, \emph{negative parameter interference} remains a challenge, since the merged model often falls short of each fine-tuned expert on its corresponding task. Such interference is exacerbated when the number of models being merged increases~\cite{ilharco2023_task-arithmetic, yadav2023tiesmerging, davari2024breadcrumbs, marczak2025isoc} or when the fine-tuning stage is prolonged~\cite{zhou2025on-task-vectors-gradients,horoi2025less-is-more}.

Many methods have been proposed to reduce negative task interference during merging and narrow the gap between the merged model and the individual experts~\cite{ilharco2023_task-arithmetic,yadav2023tiesmerging,yu2024dare,davari2024breadcrumbs,Gargiulo_2025_TSV-Merging,marczak2025isoc,yang2026model_merging}. For example, some methods rely on sparsifying the task vectors to reduce direct parameter conflicts~\cite{yadav2023tiesmerging,yu2024dare,davari2024breadcrumbs}, while others use SVD to reduce conflicts in the spectral domain of task vectors~\cite{Gargiulo_2025_TSV-Merging,marczak2025isoc}. In this work we focus on five popular and state-of-the-art merging methods, representative of the two main mechanisms used to improve task-arithmetic-based merging: sparsity and orthogonalization. We provide a brief description of each considered method below. 

\textbf{TIES} (TrIm, Elect Sign) merging was introduced in \cite{yadav2023tiesmerging} as a means to reduce parameter interference. TIES prunes the low magnitude parameters of each task vector, then averages the remaining sparse task vectors based on sign alignment: in each parameter dimension, TIES only averages the parameters from each task vector that have the same sign as the weighted majority.

\textbf{DARE} (Drop And REscale) merging~\cite{yu2024dare} randomly prunes a fraction of each task vector parameters and rescales the remaining ones based on the pruning fraction before combining them as in Eq.~\ref{eq:task_arithmetic}.

\textbf{TSV} (Task Singular Vector) merging~\cite{Gargiulo_2025_TSV-Merging} operates on the per-layer matrix form of task vectors, or task matrices, which were found empirically to be low-rank. TSV first computes a truncated SVD of each layer's task matrix to compress it into its principal singular directions, then decorrelates the singular vectors across tasks via a whitening transformation to reduce task interference before recombining.

\textbf{Iso-C} (Isotropic merging in Common subspace) \textbf{\& Iso-CTS}~\cite{marczak2025isoc} both operate on task matrices. Iso-C sums the task matrices, computes the SVD, and replaces the singular-value spectrum with its average, flattening high-energy directions to yield a more balanced common subspace. Iso-CTS further adds task-specific directions from the orthogonal complement to preserve information that may otherwise be lost during merging.



\subsection{DiLoCo as Iterative Model Merging}\label{ss:direct_analogy}
%
We begin by highlighting the structural similarity between DiLoCo and task-arithmetic-based model merging. At outer step $t$, all DiLoCo workers are initialized from the shared parameters $\theta^{(t-1)}$ and independently perform local training, producing updated parameters $\{\theta_i^{(t)}\}_{i=1}^{R}$.
DiLoCo then forms its outer update from the average of the resulting local updates, i.e., the displacement in parameter space between the shared initialization and the locally updated replicas.
Defining the local update of worker $i$ at outer step $t$ as $\Delta_i^{(t)} := \theta_i^{(t)} - \theta^{(t-1)}$, with $\Delta^{(t)}$ denoting the average local update across workers, an SGD outer optimizer with learning rate $\eta$ gives
\begin{align}
    \theta^{(t)}
    &= \theta^{(t-1)} + \eta \Delta^{(t)} \nonumber \\
    &= \theta^{(t-1)}
    + \eta \frac{1}{R}\sum_{i=1}^{R}
    \left(\theta_i^{(t)} - \theta^{(t-1)}\right).
    \label{eq:diloco_task_arithmetic}
\end{align}
%
Comparing this update to \Cref{eq:task_arithmetic} reveals a direct correspondence: the shared parameters $\theta^{(t-1)}$ play the role of the pretrained model $\theta_0$, each worker output $\theta_i^{(t)}$ plays the role of a task-specific fine-tuned model, the local updates $\Delta_i^{(t)}$ correspond to task vectors, and the number of workers corresponds to the number of tasks. The outer learning rate $\eta$ then plays the role of the task-arithmetic scaling coefficient, up to the normalization by $R$.

\textbf{Thus, with an SGD outer optimizer, DiLoCo's outer step can be viewed as an application of task-arithmetic-based model merging.} This perspective raises a natural question: \textit{can more sophisticated merging methods be substituted into the outer step to improve low-communication distributed training?} This is the question we investigate in the following section.

\section{Can Model Merging Improve Pseudo-Gradient Aggregation in DiLoCo?}\label{s:merging_improves_diloco}
\subsection{Experimental Set-up}\label{ss:setup}
\textbf{Models, data and optimizers:}
We train Llama-style decoder-only transformer models of various sizes on the DCLM~\cite{li2024dclm} dataset. The data is tokenized with the Llama-2~\citep{touvron2023llama2openfoundation} tokenizer. We use a Chinchilla budget of 20x the number of training tokens relative to the number of model parameters~\cite{hoffmann2022trainingcomputeoptimallargelanguage} and a sequence length of 2048. We use a global batch size of 524k tokens for the 178M models and 2M tokens for the 512M and 1B models. We follow the convention of matching the global batch size for our DiLoCo experiments~\cite{douillard2024diloco,charles2026scaling_laws_diloco}, i.e. for any distributed training experiment with $R$ workers, the inner batch size for each worker is set to global batch size divided by $R$. 

For all DiLoCo experiments, we fix the number of warm-up steps to 10\% of the entire training budget, we use AdamW as the inner optimizer with standard values of $\beta_1=0.9$ and $\beta_2=0.95$ and a fixed weight decay of 0.1.
Unless otherwise specified, we use $R=8$ workers, $H=30$ inner steps, and SGD with Nesterov momentum~\cite{douillard2024diloco,charles2026scaling_laws_diloco} for the outer optimizer. 

\textbf{Hyperparameter sweeps:}
For our distributed training experiments, we conduct an extensive sweep over outer momentum, outer learning rate, and inner learning rate at the 178M scale. At all other scales, we fix the outer hyperparameters (momentum and learning rate) to their optimal values at 178M and sweep only over the inner learning rate, since prior work has shown that the optimal outer hyperparameters remain approximately constant across model scales~\cite{charles2026scaling_laws_diloco,therien2026muloco}.
%
%
The inner learning rate is swept over powers of $\sqrt{2}$, while the outer momentum is swept over values in $\{0.6, 0.7, 0.8, 0.9\}$. The outer learning rate is swept from 0.5 to 2 using increments of 0.1, following past works. However, since the merging inspired outer steps can differ substantially from the standard Nesterov SGD outer optimizer, we also find it necessary to sweep values greater than 2 using an approximately logarithmic spacing to preserve the granularity of the sweep across different scales.
The training runs are evaluated on a held-out validation dataset consisting of a 100M token subset of the DCLM dataset, and the best validation loss is reported.
For the AdamW data-parallel baselines we sweep over the learning rate first with a fixed weight decay of 0.1 and then we sweep over the weight decay.

\textbf{Code and compute:}
We use the PyTorch~\cite{NEURIPS2019_pytorch} implementation from \cite{sarfi2025sparseloco}, adapted to our setting. Most of our experiments were conducted on nodes with 4xH100 Nvidia GPUs, with some of the larger scale experiments (larger $R$ and/or model size) conducted on nodes with 8xH200 Nvidia GPUs.

\subsection{Model Merging for DiLoCo Pseudo-Gradient Aggregation}
Motivated by the direct analogy between task-arithmetic-based model merging and the outer steps of DiLoCo with SGD identified in \Cref{ss:direct_analogy}, we evaluate whether other model merging methods can improve DiLoCo by replacing its outer update. We use SGD as the outer optimizer because this is the setting in which the analogy holds, and to avoid the potentially confounding effects of outer momentum. We also include data-parallel (DP) AdamW and DiLoCo with SGD outer optimizer results for comparison. Results for $R=8$ workers and $H=30$ inner steps are presented in \Cref{tab:benchmarking_merging_methods}.

\begin{wraptable}{r}{0.38\textwidth}
\vspace{-1.3em}
\centering
\small
\caption{Validation loss for DiLoCo SGD with merging-based aggregation.}
\label{tab:benchmarking_merging_methods}
\begin{tabular}{lc}
\toprule
\textbf{Method} & \textbf{Valid loss} \\
\midrule
\multicolumn{2}{l}{\textit{Baselines}} \\
AdamW DP   & 2.881 \\
DiLoCo SGD & 2.990 \\
\midrule
\multicolumn{2}{l}{\textit{DiLoCo SGD + Merging Agg.}} \\
TIES    & 2.997 \\
DARE    & 2.993 \\
TSV-M     & 3.160 \\
Iso-C   & \textbf{2.929} \\
Iso-CTS & 2.986 \\
\bottomrule
\end{tabular}
\vspace{-1em}
\end{wraptable}


We first observe that the sparsity-based methods TIES and DARE achieve final validation losses close to DiLoCo SGD, with losses of 2.997 and 2.993 respectively. However, these results occur at the lowest sparsity value tested for both methods (0.1) and performance degrades progressively as sparsity increases. This suggests that their strong performance at low sparsity is primarily due to the resulting updates being similar to those of DiLoCo SGD, rather than to the benefits of sparsification itself. Thus, sparsity does not appear to provide a useful perspective for improving DiLoCo outer optimization in this setting.

For the orthogonalization-based methods, TSV-M performs poorly, reaching a validation loss of 3.160, approximately 5.6\% higher than the DiLoCo SGD baseline. It is also among the most computationally expensive methods evaluated, requiring $T+2$ singular value decompositions for every parameter matrix in the model: one for each worker for the low-rank approximation and two additional global. While this cost may be acceptable in merging settings, where the method is only applied once after training, it becomes prohibitive in the DiLoCo setting, where the procedure is applied at every outer optimization step. We therefore exclude TSV-M from further analysis.


The related Iso-C and Iso-CTS methods both outperform the DiLoCo SGD baseline, achieving validation losses of 2.929 and 2.986 respectively. However, Iso-CTS is substantially more computationally expensive, requiring $T+1$ singular value decompositions and therefore incurring a cost on the order of the number of workers like TSV-M. Moreover, Iso-CTS performs worse than Iso-C in our setting, and prior work suggests that the difference between the two is small even for merging \citep{marczak2025isoc}. While preserving task-specific information may be marginally beneficial in model merging, where models are trained for longer and often on substantially different tasks, it appears less appropriate for DiLoCo: replicas are trained only for short inner loops, on subsets of the same dataset, and under the same objective. In this setting, the additional mechanisms introduced by Iso-CTS appear unnecessary and may even harm performance. We therefore also exclude Iso-CTS from the remainder of our analysis.

Our benchmarking of model merging methods as outer optimizers for DiLoCo therefore leaves Iso-C as the strongest candidate, improving over the DiLoCo SGD baseline by approximately 2\% with modest computational overhead: a single SVD per two-dimensional parameter matrix, rather than $O(T)$ decompositions as in TSV-M and Iso-CTS. As we will show later, even this cost can be significantly reduced. We therefore adopt Iso-C as the foundation for the remainder of our analysis.

\section{IsoLoCo: Isotropic Distributed Low-Communication Training}\label{s:isoloco}
\begin{wrapfigure}{r}{0.52\textwidth}
    \vspace{-1em}
    \centering
    \begin{minipage}{0.50\textwidth}
        \hrule height 0.8pt
        \vspace{0.5ex}
        \captionof{algorithm}{IsoLoCo}
        \label{alg:isoloco}
        \vspace{-5pt}
        \hrule height 0.4pt
        \vspace{0.75ex}
        \begin{algorithmic}[1]
            \Require Initial global model $\theta^{(0)}$
            \Require Outer learning rate $\eta_{\mathrm{out}}$, momentum $\mu$
            \State Initialize momentum buffer $B^{(0)}\gets 0$

            \For{$t=1,\ldots,T$}
                \State Complete local optimization from $\theta^{(t-1)}$
                \State Collect pseudo-gradients $\{\Delta_i^{(t)}\}_{i=1}^{R}$
                \State Compute mean $\Delta^{(t)}\gets \frac{1}{R}\sum_{i=1}^{R}\Delta_i^{(t)}$

                \For{parameter tensor $\ell$}
                    \If{$\dim(\Delta_\ell^{(t)})=2$}
                        \State $U\Sigma V^\top
                        \gets \operatorname{SVD}(\Delta_\ell^{(t)})$
                        \State $\boldsymbol{\sigma}\gets \operatorname{Diag}(\Sigma)$
                        \State $\bar{\sigma}\gets
                        \frac{1}{r}\sum_{k=1}^{r}\boldsymbol{\sigma}_{k}$
                        \State $\Delta_\ell^{(t)}\gets
                        \bar{\sigma}UV^\top$
                    \EndIf
                \EndFor

                \State $B^{(t)}\gets \mu B^{(t-1)}+\Delta^{(t)}$
                \State $P^{(t)}\gets \Delta^{(t)}+\mu B^{(t)}$
                \State $\theta^{(t)}\gets
                \theta^{(t-1)}-\eta_{\mathrm{out}}P^{(t)}$
            \EndFor
        \end{algorithmic}
        \vspace{0.75ex}
        \hrule height 0.8pt
    \end{minipage}
    \vspace{-1em}
    \vspace{-5pt}
\end{wrapfigure}

Building on Iso-C, we introduce IsoLoCo (\textbf{Iso}tropic distributed \textbf{Lo}w-\textbf{Co}mmunication) training, which augments the isotropic correction with Nesterov momentum. This is motivated by \cite{douillard2024diloco}, who identified Nesterov momentum as highly beneficial for FedOpt algorithms in distributed low-communication LLM training; we hypothesize it can also improve Iso-C.

The algorithm is presented in Algorithm~\ref{alg:isoloco}. At each outer step $t$, each worker $i$ provides a pseudo-gradient $\Delta_i^{(t)} := \theta_i^{(t)} - \theta^{(t-1)}$, representing its local progress relative to the previous synchronized global parameters $\theta^{(t-1)}$. These are averaged to form $\Delta^{(t)}$. For weight matrices (parameters with exactly two dimensions), we apply the isotropic correction: the SVD of $\Delta^{(t)}$ is computed, its singular values replaced by their mean $\bar{\sigma}$, and the update reconstructed as $\bar{\sigma} U V^\top$, projecting onto a nearby isotropic operator that regularizes the spectrum while preserving subspace structure. Non-matrix parameters (e.g.\ normalization layers) are left unchanged. The corrected pseudo-gradient is then incorporated into a Nesterov momentum scheme: the buffer $B^{(t)}$ accumulates past corrected pseudo-gradients, and the effective update $P^{(t)}$ applies the characteristic look-ahead.

\subsection{IsoLoCo Outperforms DiLoCo at High Worker Counts}
We now review our low-communication LLM pre-training results, where the number of workers we train on grows from $R=1$ to $128$ and the synchronization interval is fixed at $H=30$. 
To the best of our knowledge, we scaled the number of workers well beyond the largest settings studied in past works $R\approx16$--$20$~\cite{therien2026muloco,lidin2026covenant72bpretraining72bllm,intellect1}. 
Our results are presented in \Cref{tab:valid-loss-workers,tab:1b_models} and \Cref{fig:workerscaling}. 

\Cref{tab:valid-loss-workers} presents the final validation loss of DiLoCo, DiLoCo SGD, DiLoCo SGD + Iso-C, and IsoLoCo on 178M models and, for DiLoCo and IsoLoCo, on 512M models. Surprisingly, even the naive combination of DiLoCo SGD with Iso-C merging-based aggregation outperforms DiLoCo at 178M scale across all worker counts, despite lacking a momentum mechanism. IsoLoCo further widens this gap, outperforming DiLoCo across both model scales and confirming the added benefit of momentum. At 178M scale, we find that the improvement of $0.027$ loss at $R=1$ ($\sim0.9\%$) grows to $0.195$ at $R=128$ ($\sim5.8\%$), showing IsoLoCo's substantial improvements over DiLoCo at large worker counts.
At the 512M scale, IsoLoCo's improvement over DiLoCo is just as strong, growing from $0.041$ ($1.6\%$) at $R=1$ to $0.147$ ($5.1\%$) at $R=64$.
The $R=128$ results for the 512M models are omitted due to out-of-memory issues at that scale.

In \Cref{tab:1b_models} we also report results for 1B-scale models with $R=8$ and $H=30$. IsoLoCo continues to outperform DiLoCo at this scale ($2.421$ vs.\ $2.450$), though the relative improvement is more modest than at smaller scales: $1.2\%$ at 1B, compared to $2.1\%$ and $2.4\%$ at 178M and 512M (all with $R=8$). Two effects help explain this. First, DiLoCo's worker scaling is known to improve with model size~\cite{charles2026scaling_laws_diloco,therien2026muloco}, allowing it to catch up to IsoLoCo at the larger scales. Second, larger models reach lower validation loss overall, leaving less headroom for algorithmic gains. Taken together, these suggest that IsoLoCo should continue to outperform DiLoCo as model size increases, albeit with smaller relative improvements at larger scales.

\begin{table}[t]
\centering
\small
\caption{\textbf{Validation loss across numbers of workers $R$ and model sizes.} The AdamW DP baseline reached $2.881$ and $2.633$ final loss at 178M and 512M scales, respectively. We observe that IsoLoCo consistently improves over DiLoCo and has much stronger performance at large worker counts. }
\label{tab:valid-loss-workers}
\begin{adjustbox}{width=\linewidth}
\begin{tabular}{llcccccccc}
\toprule
\textbf{Model} & \textbf{Method} & \textbf{$R=1$} & \textbf{$R=2$} & \textbf{$R=4$} & \textbf{$R=8$} & \textbf{$R=16$} & \textbf{$R=32$} & \textbf{$R=64$} & \textbf{$R=128$} \\
\midrule
\multirow{4}{*}{178M}
& DiLoCo SGD           & 2.887 & 2.905 & 2.937 & 2.990 & 3.064 & 3.178 & 3.357 & 3.601 \\
& DiLoCo               & 2.869 & 2.882 & 2.911 & 2.950 & 3.005 & 3.077 & 3.195 & 3.355 \\
& DiLoCo SGD + Iso-C   & 2.855 & 2.872 & 2.891 & 2.929 & 2.992 & 3.070 & 3.162 & 3.280 \\
&\cellcolor{gray!15}IsoLoCo (ours)       &\cellcolor{gray!15}{\bf 2.842} &\cellcolor{gray!15}{\bf 2.852} &\cellcolor{gray!15}{\bf 2.864} &\cellcolor{gray!15}{\bf 2.889} &\cellcolor{gray!15}{\bf 2.930} &\cellcolor{gray!15}{\bf 2.992} &\cellcolor{gray!15}{\bf 3.062} &\cellcolor{gray!15}{\bf 3.160} \\
\midrule
\multirow{2}{*}{512M}
& DiLoCo          & 2.630 & 2.639 & 2.653 & 2.675 & 2.706 & 2.767 & 2.863 & -- \\
&\cellcolor{gray!15}IsoLoCo (ours)  &\cellcolor{gray!15}{\bf 2.589} &\cellcolor{gray!15}{\bf 2.593} &\cellcolor{gray!15}{\bf 2.599} &\cellcolor{gray!15}{\bf 2.610} &\cellcolor{gray!15}{\bf 2.628} &\cellcolor{gray!15}{\bf 2.676} &\cellcolor{gray!15}{\bf 2.716} &\cellcolor{gray!15}-- \\
\bottomrule
\end{tabular}
\end{adjustbox}
\vspace{-10pt}
\end{table}

\begin{wraptable}{r}{0.38\textwidth}
\centering
\small
\caption{Validation loss for DiLoCo and IsoLoCo for 1B models, ${R=8}$.}
\label{tab:1b_models}
\begin{tabular}{lc}
\toprule
\textbf{Method} & \textbf{Valid loss} \\
DiLoCo    & 2.450\\
\rowcolor{gray!15}IsoLoCo (ours)   & \textbf{2.421} \\
\bottomrule
\end{tabular}
\vspace{-1em}
\end{wraptable}
\Cref{fig:workerscaling} reports the performance of DiLoCo, DiLoCo SGD + Iso-C, and IsoLoCo at 178M scale normalized relative to an AdamW DP baseline. Specifically, we report the percentage $100\cdot(\frac{\text{Method}}{\text{AdamW DP}}-1)$, i.e. the relative increase in loss over AdamW DP. Since communication-intensive data-parallel training is the de facto gold standard and low communication methods are known to degrade in performance at larger worker counts~\cite{therien2026muloco, charles2026scaling_laws_diloco}, this metric allows direct inspection of the performance degradation of different methods relative to this ideal. We observe that relative to the AdamW DP (Loss=$2.881$) all low communication methods degrade as the number of workers increases. However, IsoLoCo strictly improves over DiLoCo at all worker counts, reaching only a 9.7\% degradation at $R=128$ vs. DiLoCo's 16.5\%. Moreover, IsoLoCo outperforms AdamW DP at $R\leq4$. In summary, we find that IsoLoCo yields strict improvements over DiLoCo in all settings.

\subsection{Increasing the Number of Inner Steps}
We now evaluate the performance of our method as the synchronization interval is increased from $H=30$ to $240$. The number of workers is fixed to $R=8$ here.
\Cref{tab:inner-steps-scaling} reports the final loss of DiLoCo and IsoLoCo at each synchronization interval $H\in\{30, 60, 120, 240\}$. Both methods were extensively tuned in each case. We observe that IsoLoCo consistently improves over DiLoCo across all numbers of inner steps. These results demonstrate that IsoLoCo is compatible with long synchronization intervals, which is essential for training under heavy bandwidth constraints.

\begin{table}[ht]
\vspace{-7pt}
\centering
\caption{\textbf{Validation loss as a function of synchronization interval ($H$) for DiLoCo and IsoLoCo.} We observe that IsoLoCo consistently improves over DiLoCo across all synchronization intervals.  }
\label{tab:inner-steps-scaling}
\begin{tabular}{lcccc}
\toprule
\textbf{Method} & \textbf{$H=30$} & \textbf{$H=60$} & \textbf{$H=120$} & \textbf{$H=240$} \\
\midrule
DiLoCo & 2.950 & 2.957 & 2.976 & 3.008 \\
\rowcolor{gray!15}IsoLoCo (ours) & {\bf2.889} & {\bf2.907} & {\bf2.944} & {\bf2.991} \\
\bottomrule
\end{tabular}
\vspace{-5pt}
\end{table}

\subsection{Accelerating IsoLoCo with Newton-Schulz Iterations}\label{ss:fast_isoloco}
Despite being significantly less computationally expensive than some of the other merging methods considered, Iso-C still relies on SVD to compute the singular values and vectors of the gradients, which can be slow on GPUs. We therefore explore whether IsoLoCo can be accelerated using the iterative Newton-Schulz~\cite{bjork1971iterative_algorithm,kovarik1970iterative_methods} method, which is already used in the recently popularized Muon optimizer~\cite{jordan2024muon,liu2025muonscalablellmtraining}. We discuss the relationship between IsoLoCo and Muon in more detail in \Cref{a:iso_vs_muon}.
Let ${M}\in\mathbb{R}^{n\times m}$ be a matrix with singular value decomposition ${M}={U S V^\top}$. The Newton-Schulz ($\operatorname{NS}$) method uses matrix multiplications to approximately orthogonalize ${M}$, i.e.\ $\operatorname{NS}({M})\approx {UV^\top}$.
However, using Newton-Schulz to orthogonalize the gradients does not yield the individual singular values, which are needed to compute their mean in IsoLoCo. Instead, we propose using the root mean square of the singular values as a cheap proxy, since it can be computed directly from the gradient matrix before orthogonalization:
\[
    \operatorname{RMS}(\sigma)=\frac{\|{M}\|_F}{\sqrt{r}}= \sqrt{\frac{1}{r}\sum\limits_{i=1}^r \sigma^2_i}
\]
Note that in general $\operatorname{RMS}(\sigma)\geq\operatorname{mean}(\sigma)$, with equality only when all singular values are identical.
Despite this difference, we find that the $\operatorname{RMS}$ is a good replacement for the mean. In our experiments with 178M models, $R=8$, and $H=30$, the fast implementation reaches a validation loss of 2.902, still significantly outperforming DiLoCo (2.950) and trailing the SVD implementation (2.889) by only ${\sim}0.45\%$. Furthermore, both versions share the same optimal hyperparameters.

\textbf{Time complexity analysis.}
Both SVD and Newton-Schulz have a time complexity of $O(\min(m^2n, mn^2))$, with Newton-Schulz also depending on the number of iterations; however, that number is small (we use the standard 5-step iteration from the literature~\cite{jordan2024muon,liu2025muonscalablellmtraining} with the polynomial coefficients from \cite{ahn2025diondistributedorthonormalizedupdates}). In practice, Newton-Schulz is faster because it relies solely on general matrix-matrix multiplications, which GPUs are heavily optimized for and which can exploit lower-precision arithmetic. 
At the 178M model scale with $R=8$, we found the Newton--Schulz orthogonalization step to be \textbf{up to two orders of magnitude faster} than SVD for the matrix sizes used in our models, resulting in a roughly 15\% speed-up in wall-clock time over the entire training duration, with all other factors held constant.

\subsection{IsoLoCo improves MuLoCo's worker-count scaling}\label{s:muon_inner}
The Muon optimizer~\cite{jordan2024muon} has recently attracted considerable attention for LLM pre-training~\cite{liu2025muonscalablellmtraining,ai2025practicalefficiencymuonpretraining}, as it improves on both the performance and efficiency of the long-standing default optimizer, AdamW~\cite{loshchilov2018adamw}. Closest to our setting, \cite{therien2026muloco} use Muon as DiLoCo's \emph{inner} optimizer and show that the resulting method, MuLoCo, scales better with worker count, tolerates tighter communication constraints, and admits larger critical batch sizes than standard DiLoCo. Since IsoLoCo modifies only the \emph{outer} step, it is orthogonal to the choice of inner optimizer. We therefore ask whether the two improvements compose, pairing IsoLoCo's outer step with a Muon inner optimizer.

\begin{wrapfigure}{r}{0.55\linewidth}
  \centering
  \vspace{-12pt}
  \includegraphics[width=\linewidth]{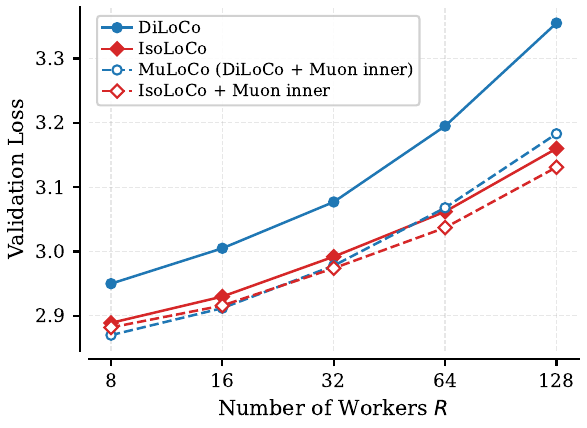} \vspace{-15pt}
  \caption{\textbf{IsoLoCo further improves MuLoCo's worker-count scaling.} Final validation loss is shown versus worker count for a 178M LLaMa-2 model in a FLOP-matched setting.
  \vspace{-10pt}
  }
  \label{fig:muon_inner}
\end{wrapfigure}

\Cref{fig:muon_inner} compares DiLoCo and IsoLoCo under both AdamW~\cite{loshchilov2018adamw} and Muon~\cite{jordan2024muon} inner optimizers as the worker count is varied from $R=8$ to $R=128$ in a FLOP-matched setting; following \cite{therien2026muloco}, we denote DiLoCo with a Muon inner optimizer as MuLoCo.

The two techniques improve DiLoCo through the same underlying mechanism---flattening the spectrum of the updates---but at different stages: IsoLoCo replaces the singular values of the outer update with their mean, while Muon drives the singular values of the inner updates toward one. This shared mechanism explains both why combining them helps and why the effect is not additive. Each operation improves over DiLoCo on its own, and because they act at different stages they contribute complementary gains: IsoLoCo's isotropic scaling improves the aggregation of worker gradients, yielding its largest advantages at high worker counts, while Muon's inner steps improve convergence. Combining them therefore helps, and increasingly so as workers scale. But since both operations flatten the spectrum, stacking them does not double the benefit. Because the outer step is applied only once every $H$ inner steps, IsoLoCo orthogonalizes far less frequently than MuLoCo, which does so at every inner step. IsoLoCo is therefore considerably cheaper while matching MuLoCo at low worker counts and surpassing it at high ones.

This is borne out by the numbers. At $R=8$, IsoLoCo, MuLoCo, and IsoLoCo with a Muon inner optimizer all fall within ${\sim}0.7\%$ of one another, and all sit more than ${\sim}2\%$ below DiLoCo: at low worker counts the two operations do largely the same job, so little is gained by applying both. As the worker count grows, however, the combination of IsoLoCo and MuLoCo increasingly separates from either method alone. For $R\geq32$ it attains the lowest validation loss among all methods considered, with its improvement over MuLoCo growing from $0.13\%$ at $R=32$ to $1.63\%$ at $R=128$, and its improvement over IsoLoCo with an AdamW inner optimizer growing from $0.60\%$ to $0.92\%$ over the same range. This is consistent with the outer and inner interventions addressing distinct bottlenecks, namely gradient aggregation and convergence respectively, that both become more pronounced at scale.
%

\subsection{Ablations and Analysis}\label{ss:ablations}
We conduct ablations on our 178M models with $R=8$ to validate key design choices in IsoLoCo.

\textbf{Momentum and orthogonalization target.}
We compare Nesterov momentum against standard (non-Nesterov) momentum and evaluate whether orthogonalization should be applied to the gradients $\Delta^{(t)}$ (as in IsoLoCo) or to the final update matrix $P^{(t)}$. For the momentum experiments, we additionally sweep over the outer momentum coefficient, as the non-Nesterov variant requires lower values for this hyperparameter. We find that all design choices have a modest effect: non-Nesterov momentum yields slightly worse results (${\sim}0.2\%$) than the Nesterov variant, and applying orthogonalization to $P^{(t)}$ rather than the gradients also degrades performance slightly. We additionally tried excluding the two-dimensional embedding and head parameters from orthogonalization but observed no improvement.

\textbf{Directional analysis of singular value equalization.}
Since IsoLoCo exposes the gradients' singular values, we can isolate the source of its gains: does performance improve primarily from clipping high singular values down to the mean, or from boosting low singular values up to it? We implement two variants: \textsc{clip\_high}, which sets only above-mean singular values to the mean, and \textsc{clip\_low}, which does the same for below-mean values, leaving the remaining singular values unchanged.
Both variants improve over the DiLoCo baseline (2.950), with \textsc{clip\_low} reaching 2.931 and \textsc{clip\_high} reaching 2.894. Notably, \textsc{clip\_high} nearly matches the full IsoLoCo result (2.889), while \textsc{clip\_low} closes only a small fraction of the gap.
\textbf{We therefore conclude that IsoLoCo's gains are primarily driven by reducing the dominance of a few high-energy directions in the gradient spectrum.}

\section{Conclusion and Limitations}\label{s:conclusion_limitations}
This work investigates model merging as a mechanism for improving gradient aggregation in DiLoCo-style low-communication training, providing what is, to our knowledge, the first thorough study of model merging methods in the context of distributed optimization. Our analysis is grounded in the direct analogy between task-arithmetic-based model merging and DiLoCo with SGD outer optimizer (\Cref{ss:direct_analogy}). Evaluating merging methods in this setting allows us to identify Iso-C as a particularly effective replacement for pseudo-gradient averaging and propose IsoLoCo, a momentum-equipped variant for distributed optimization. IsoLoCo consistently outperforms DiLoCo across model sizes and synchronization intervals, with gains that become more pronounced as the number of workers increases. We further introduce a fast implementation based on the iterative Newton--Schulz method that closely matches the SVD version while yielding substantial wall-clock improvements.

More broadly, our results suggest that model merging is a valuable source of inspiration for distributed training research: merging-inspired gradient aggregation can help low-communication training scale to larger numbers of poorly interconnected workers while reducing the performance gap to AdamW DP typically observed with standard DiLoCo. Important future directions include evaluating IsoLoCo under strong compression, combining it with low-communication pipelining methods, studying the single-worker case more extensively, and developing a theoretical understanding of why merging-inspired distributed optimizers, IsoLoCo in particular, improve results.

\textbf{Limitations.} 
While we have made our best effort to provide a thorough scientific study and taken the necessary measures to ensure our results are sound, limitations naturally arise from the computational resources available in academia. In no particular order, our empirical results do not evaluate models beyond 1B scale, architectures beyond a dense transformer (e.g.\ MoE, efficient attention variants, hybrid architectures), or tasks beyond language modeling. On the theoretical side, we do not prove convergence for optimizers employing model merging outer steps.

\textbf{Broader Impact.} This work proposes improvements to distributed optimization for neural network training, which could help make large-scale training more accessible to researchers with limited or heterogeneous infrastructure. We do not foresee specific negative societal consequences beyond those generally associated with advances in language modeling.


\section*{Acknowledgments}
This work was partially funded by the NSERC CGS D 569345 - 2022 scholarship [S.H.]; FRQNT Doctoral (B2X) scholarship [B.T.]; FRQNT New Scholar grant [E.B.]; FRQNT-NSERC grant 2023-NOVA-329125 [E.B. \& G.W.]; Canada CIFAR AI Chair, NSF DMS grant 2327211 and NSERC Discovery grant 03267 [G.W.]. 
This work is also supported by resources from Compute Canada and Calcul Quebec.
The content is solely the responsibility of the authors and does not necessarily represent the views of the funding agencies.


\newpage
\bibliography{main}

@misc{douillard2024diloco,
      title={DiLoCo: Distributed Low-Communication Training of Language Models}, 
      author={Arthur Douillard and Qixuan Feng and Andrei A. Rusu and Rachita Chhaparia and Yani Donchev and Adhiguna Kuncoro and Marc'Aurelio Ranzato and Arthur Szlam and Jiajun Shen},
      year={2024},
      eprint={2311.08105},
      archivePrefix={arXiv},
      primaryClass={cs.LG},
      url={https://arxiv.org/abs/2311.08105}, 
}

@inproceedings{
charles2026scaling_laws_diloco,
title={Communication-Efficient Language Model Training Scales Reliably and Robustly: Scaling Laws for DiLoCo},
author={Zachary Charles and Gabriel Teston and Lucio M. Dery and J Keith Rush and Nova Fallen and Zachary Garrett and Arthur Szlam and Arthur Douillard},
booktitle={The Thirty-ninth Annual Conference on Neural Information Processing Systems},
year={2026},
url={https://openreview.net/forum?id=X4SCxcgb3O}
}

@InProceedings{pmlr-v54-mcmahan17a-fedavg,
  title = 	 {{Communication-Efficient Learning of Deep Networks from Decentralized Data}},
  author = 	 {McMahan, Brendan and Moore, Eider and Ramage, Daniel and Hampson, Seth and Arcas, Blaise Aguera y},
  booktitle = 	 {Proceedings of the 20th International Conference on Artificial Intelligence and Statistics},
  pages = 	 {1273--1282},
  year = 	 {2017},
  editor = 	 {Singh, Aarti and Zhu, Jerry},
  volume = 	 {54},
  series = 	 {Proceedings of Machine Learning Research},
  month = 	 {20--22 Apr},
  publisher =    {PMLR},
  url = 	 {https://proceedings.mlr.press/v54/mcmahan17a.html}
}

@inproceedings{
reddi2021fedopt,
title={Adaptive Federated Optimization},
author={Sashank J. Reddi and Zachary Charles and Manzil Zaheer and Zachary Garrett and Keith Rush and Jakub Kone{\v{c}}n{\'y} and Sanjiv Kumar and Hugh Brendan McMahan},
booktitle={International Conference on Learning Representations},
year={2021},
url={https://openreview.net/forum?id=LkFG3lB13U5}
}

@InProceedings{kingma2015adammethodstochasticoptimization,
      title={Adam: A Method for Stochastic Optimization}, 
      author={Diederik P. Kingma and Jimmy Ba},
      year={2015},
      url={https://arxiv.org/abs/1412.6980},
  booktitle     = {International Conference on Learning Representations~(ICLR)}
}

@inproceedings{
loshchilov2018adamw,
title={Decoupled Weight Decay Regularization},
author={Ilya Loshchilov and Frank Hutter},
booktitle={International Conference on Learning Representations~(ICLR)},
year={2019},
url={https://openreview.net/forum?id=Bkg6RiCqY7},
}

@InProceedings{sutskever2013init_momentum,
  title = 	 {On the importance of initialization and momentum in deep learning},
  author = 	 {Sutskever, Ilya and Martens, James and Dahl, George and Hinton, Geoffrey},
  booktitle = 	 {International Conference on Machine Learning~(ICML)},
  pages = 	 {1139--1147},
  year = 	 {2013},
  editor = 	 {Dasgupta, Sanjoy and McAllester, David},
  volume = 	 {28},
  series = 	 {Proceedings of Machine Learning Research},
  address = 	 {Atlanta, Georgia, USA},
  month = 	 {17--19 Jun},
  publisher =    {PMLR},
  url = 	 {https://proceedings.mlr.press/v28/sutskever13.html}
}

@article{Nesterov1983AMF,
  title={A method for solving the convex programming problem with convergence rate {$O(1/k^2)$}},
  author={Yurii Nesterov},
  journal={Proceedings of the USSR Academy of Sciences},
  year={1983},
  volume={269},
  pages={543-547},
  url={https://api.semanticscholar.org/CorpusID:145918791}
}

@inproceedings{garipov2018_fge,  
author = {Garipov, Timur and Izmailov, Pavel and Podoprikhin, Dmitrii and Vetrov, Dmitry P and Wilson, Andrew G},  
booktitle = {Advances in Neural Information Processing Systems~(NeurIPS)},  
editor = {S. Bengio and H. Wallach and H. Larochelle and K. Grauman and N. Cesa-Bianchi and R. Garnett},  
pages = {},  
publisher = {Curran Associates, Inc.},  
title = {Loss Surfaces, Mode Connectivity, and Fast Ensembling of {DNN}s},  
url = {https://proceedings.neurips.cc/paper_files/paper/2018/file/be3087e74e9100d4bc4c6268cdbe8456-Paper.pdf},  
volume = {31},  
year = {2018}  
}

@InProceedings{draxler2018_no-barriers,  
title = {Essentially No Barriers in Neural Network Energy Landscape},  
author = {Draxler, Felix and Veschgini, Kambis and Salmhofer, Manfred and Hamprecht, Fred},  
booktitle = {International Conference on Machine Learning~(ICML)},  
pages = {1309--1318},  
year = {2018},  
editor = {Dy, Jennifer and Krause, Andreas},  
volume = {80},  
series = {Proceedings of Machine Learning Research},  
month = {10--15 Jul},  
publisher = {PMLR},  
url = {https://proceedings.mlr.press/v80/draxler18a.html}
}

@InProceedings{pmlr-v119-frankle20a_lmc_lth,  
title = {Linear Mode Connectivity and the Lottery Ticket Hypothesis},  
author = {Frankle, Jonathan and Dziugaite, Gintare Karolina and Roy, Daniel and Carbin, Michael},  
booktitle = {International Conference on Machine Learning~(ICML)},  
pages = {3259--3269},  
year = {2020},  
editor = {III, Hal Daumé and Singh, Aarti},  
volume = {119},  
series = {Proceedings of Machine Learning Research},  
month = {13--18 Jul},  
publisher = {PMLR},  
url = {https://proceedings.mlr.press/v119/frankle20a.html}
}

@inproceedings{neyshabur2020transfer,
author = {Neyshabur, Behnam and Sedghi, Hanie and Zhang, Chiyuan},
booktitle = {Advances in Neural Information Processing Systems~(NeurIPS)},
editor = {H. Larochelle and M. Ranzato and R. Hadsell and M.F. Balcan and H. Lin},
pages = {512--523},
publisher = {Curran Associates, Inc.},
title = {What is being transferred in transfer learning? },
url = {https://proceedings.neurips.cc/paper/2020/file/0607f4c705595b911a4f3e7a127b44e0-Paper.pdf},
volume = {33},
year = {2020}
}

@inproceedings{
ilharco2023_task-arithmetic,
title={Editing models with task arithmetic},
author={Gabriel Ilharco and Marco Tulio Ribeiro and Mitchell Wortsman and Ludwig Schmidt and Hannaneh Hajishirzi and Ali Farhadi},
booktitle={International Conference on Learning Representations~(ICLR)},
year={2023},
url={https://openreview.net/forum?id=6t0Kwf8-jrj}
}

@inproceedings{yadav2023tiesmerging,
 author = {Yadav, Prateek and Tam, Derek and Choshen, Leshem and Raffel, Colin A and Bansal, Mohit},
 booktitle = {Advances in Neural Information Processing Systems~(NeurIPS)},
 editor = {A. Oh and T. Naumann and A. Globerson and K. Saenko and M. Hardt and S. Levine},
 pages = {7093--7115},
 publisher = {Curran Associates, Inc.},
 title = {TIES-Merging: Resolving Interference When Merging Models},
 url = {https://proceedings.neurips.cc/paper_files/paper/2023/file/1644c9af28ab7916874f6fd6228a9bcf-Paper-Conference.pdf},
 volume = {36},
 year = {2023}
}

@InProceedings{yu2024dare,
title = {Language Models are Super Mario: Absorbing Abilities from Homologous Models as a Free Lunch},
author = {Yu, Le and Yu, Bowen and Yu, Haiyang and Huang, Fei and Li, Yongbin},
booktitle = {International Conference on Machine Learning~(ICML)},
pages = {57755--57775},
year = {2024},
editor = {Salakhutdinov, Ruslan and Kolter, Zico and Heller, Katherine and Weller, Adrian and Oliver, Nuria and Scarlett, Jonathan and Berkenkamp, Felix},
volume = {235},
series = {Proceedings of Machine Learning Research},
month = {21--27 Jul},
publisher = {PMLR},
url = {https://proceedings.mlr.press/v235/yu24p.html},
}

@InProceedings{Gargiulo_2025_TSV-Merging,
    author    = {Gargiulo, Antonio Andrea and Crisostomi, Donato and Bucarelli, Maria Sofia and Scardapane, Simone and Silvestri, Fabrizio and Rodol\`a, Emanuele},
    title     = {Task Singular Vectors: Reducing Task Interference in Model Merging},
    booktitle = {Proceedings of the IEEE/CVF Conference on Computer Vision and Pattern Recognition (CVPR)},
    month     = {June},
    year      = {2025},
    pages     = {18695-18705}
}

@inproceedings{
marczak2025isoc,
title={No Task Left Behind: Isotropic Model Merging with Common and Task-Specific Subspaces},
author={Daniel Marczak and Simone Magistri and Sebastian Cygert and Bart{\l}omiej Twardowski and Andrew D. Bagdanov and Joost van de Weijer},
booktitle={Forty-second International Conference on Machine Learning},
year={2025},
url={https://openreview.net/forum?id=RBZpAa27ls}
}

@misc{jordan2024muon,
  author       = {Jordan, Keller and Jin, Yuchen and Boza, Vlado and You, Jiacheng and
                  Cesista, Franz and Newhouse, Laker and Bernstein, Jeremy},
  title        = {Muon: An optimizer for hidden layers in neural networks},
  year         = {2024},
  howpublished = {\url{https://kellerjordan.github.io/posts/muon/}},
  note         = {Accessed: 2026-04-27}
}

@misc{liu2025muonscalablellmtraining,
      title={Muon is Scalable for LLM Training}, 
      author={Jingyuan Liu and Jianlin Su and Xingcheng Yao and Zhejun Jiang and Guokun Lai and Yulun Du and Yidao Qin and Weixin Xu and Enzhe Lu and Junjie Yan and Yanru Chen and Huabin Zheng and Yibo Liu and Shaowei Liu and Bohong Yin and Weiran He and Han Zhu and Yuzhi Wang and Jianzhou Wang and Mengnan Dong and Zheng Zhang and Yongsheng Kang and Hao Zhang and Xinran Xu and Yutao Zhang and Yuxin Wu and Xinyu Zhou and Zhilin Yang},
      year={2025},
      eprint={2502.16982},
      archivePrefix={arXiv},
      primaryClass={cs.LG},
      url={https://arxiv.org/abs/2502.16982}, 
}

@misc{yang2024spectralconditionfeaturelearning,
      title={A Spectral Condition for Feature Learning}, 
      author={Greg Yang and James B. Simon and Jeremy Bernstein},
      year={2024},
      eprint={2310.17813},
      archivePrefix={arXiv},
      primaryClass={cs.LG},
      url={https://arxiv.org/abs/2310.17813}, 
}

@misc{sarfi2025sparseloco,
  title        = {Communication Efficient LLM Pre-training with SparseLoCo},
  author       = {Sarfi, Amir and Thérien, Benjamin and Lidin, Joel and Belilovsky, Eugene},
  year         = {2025},
  eprint       = {2508.15706},
  archivePrefix= {arXiv},
  primaryClass = {cs.LG},
  howpublished = {\url{https://arxiv.org/pdf/2508.15706}}
}

@inproceedings{li2024dclm,
 author = {Li, Jeffrey and Fang, Alex and Smyrnis, Georgios and Ivgi, Maor and Jordan, Matt and Gadre, Samir and Bansal, Hritik and Guha, Etash and Keh, Sedrick and Arora, Kushal and Garg, Saurabh and Xin, Rui and Muennighoff, Niklas and Heckel, Reinhard and Mercat, Jean and Chen, Mayee and Gururangan, Suchin and Wortsman, Mitchell and Albalak, Alon and Bitton, Yonatan and Nezhurina, Marianna and Abbas, Amro and Hsieh, Cheng-Yu and Ghosh, Dhruba and Gardner, Josh and Kilian, Maciej and Zhang, Hanlin and Shao, Rulin and Pratt, Sarah and Sanyal, Sunny and Ilharco, Gabriel and Daras, Giannis and Marathe, Kalyani and Gokaslan, Aaron and Zhang, Jieyu and Chandu, Khyathi and Nguyen, Thao and Vasiljevic, Igor and Kakade, Sham and Song, Shuran and Sanghavi, Sujay and Faghri, Fartash and Oh, Sewoong and Zettlemoyer, Luke and Lo, Kyle and El-Nouby, Alaaeldin and Pouransari, Hadi and Toshev, Alexander and Wang, Stephanie and Groeneveld, Dirk and Soldaini, Luca and Koh, Pang Wei and Jitsev, Jenia and Kollar, Thomas and Dimakis, Alexandros G. and Carmon, Yair and Dave, Achal and Schmidt, Ludwig and Shankar, Vaishaal},
 booktitle = {Advances in Neural Information Processing Systems},
 doi = {10.52202/079017-0455},
 editor = {A. Globerson and L. Mackey and D. Belgrave and A. Fan and U. Paquet and J. Tomczak and C. Zhang},
 pages = {14200--14282},
 publisher = {Curran Associates, Inc.},
 title = {DataComp-LM: In search of the next generation of training sets for language models},
 url = {https://proceedings.neurips.cc/paper_files/paper/2024/file/19e4ea30dded58259665db375885e412-Paper-Datasets_and_Benchmarks_Track.pdf},
 volume = {37},
 year = {2024}
}

@misc{touvron2023llama2openfoundation,
      title={Llama 2: Open Foundation and Fine-Tuned Chat Models}, 
      author={Hugo Touvron and Louis Martin and Kevin Stone and Peter Albert and Amjad Almahairi and Yasmine Babaei and Nikolay Bashlykov and Soumya Batra and Prajjwal Bhargava and Shruti Bhosale and Dan Bikel and Lukas Blecher and Cristian Canton Ferrer and Moya Chen and Guillem Cucurull and David Esiobu and Jude Fernandes and Jeremy Fu and Wenyin Fu and Brian Fuller and Cynthia Gao and Vedanuj Goswami and Naman Goyal and Anthony Hartshorn and Saghar Hosseini and Rui Hou and Hakan Inan and Marcin Kardas and Viktor Kerkez and Madian Khabsa and Isabel Kloumann and Artem Korenev and Punit Singh Koura and Marie-Anne Lachaux and Thibaut Lavril and Jenya Lee and Diana Liskovich and Yinghai Lu and Yuning Mao and Xavier Martinet and Todor Mihaylov and Pushkar Mishra and Igor Molybog and Yixin Nie and Andrew Poulton and Jeremy Reizenstein and Rashi Rungta and Kalyan Saladi and Alan Schelten and Ruan Silva and Eric Michael Smith and Ranjan Subramanian and Xiaoqing Ellen Tan and Binh Tang and Ross Taylor and Adina Williams and Jian Xiang Kuan and Puxin Xu and Zheng Yan and Iliyan Zarov and Yuchen Zhang and Angela Fan and Melanie Kambadur and Sharan Narang and Aurelien Rodriguez and Robert Stojnic and Sergey Edunov and Thomas Scialom},
      year={2023},
      eprint={2307.09288},
      archivePrefix={arXiv},
      primaryClass={cs.CL},
      url={https://arxiv.org/abs/2307.09288}, 
}

@misc{hoffmann2022trainingcomputeoptimallargelanguage,
      title={Training Compute-Optimal Large Language Models}, 
      author={Jordan Hoffmann and Sebastian Borgeaud and Arthur Mensch and Elena Buchatskaya and Trevor Cai and Eliza Rutherford and Diego de Las Casas and Lisa Anne Hendricks and Johannes Welbl and Aidan Clark and Tom Hennigan and Eric Noland and Katie Millican and George van den Driessche and Bogdan Damoc and Aurelia Guy and Simon Osindero and Karen Simonyan and Erich Elsen and Jack W. Rae and Oriol Vinyals and Laurent Sifre},
      year={2022},
      eprint={2203.15556},
      archivePrefix={arXiv},
      primaryClass={cs.CL},
      url={https://arxiv.org/abs/2203.15556}, 
}

@inproceedings{NEURIPS2019_pytorch,
 author = {Paszke, Adam and Gross, Sam and Massa, Francisco and Lerer, Adam and Bradbury, James and Chanan, Gregory and Killeen, Trevor and Lin, Zeming and Gimelshein, Natalia and Antiga, Luca and Desmaison, Alban and Kopf, Andreas and Yang, Edward and DeVito, Zachary and Raison, Martin and Tejani, Alykhan and Chilamkurthy, Sasank and Steiner, Benoit and Fang, Lu and Bai, Junjie and Chintala, Soumith},
 booktitle = {Advances in Neural Information Processing Systems~(NeurIPS)},
 editor = {H. Wallach and H. Larochelle and A. Beygelzimer and F. d\textquotesingle Alch\'{e}-Buc and E. Fox and R. Garnett},
 pages = {},
 publisher = {Curran Associates, Inc.},
 title = "{PyTorch}: An Imperative Style, High-Performance Deep Learning Library",
 url = {https://proceedings.neurips.cc/paper_files/paper/2019/file/bdbca288fee7f92f2bfa9f7012727740-Paper.pdf},
 volume = {32},
 year = {2019}
}

@misc{therien2026muloco,
      title={MuLoCo: Muon is a practical inner optimizer for DiLoCo}, 
      author={Benjamin Thérien and Xiaolong Huang and Aaron Defazio and Irina Rish and Eugene Belilovsky},
      year={2026},
      eprint={2505.23725},
      archivePrefix={arXiv},
      primaryClass={cs.LG},
      url={https://arxiv.org/abs/2505.23725}, 
}

@inproceedings{pfeiffer-etal-2020-adapterhub,
    title = "{A}dapter{H}ub: A Framework for Adapting Transformers",
    author = {Pfeiffer, Jonas  and
      R{\"u}ckl{\'e}, Andreas  and
      Poth, Clifton  and
      Kamath, Aishwarya  and
      Vuli{\'c}, Ivan  and
      Ruder, Sebastian  and
      Cho, Kyunghyun  and
      Gurevych, Iryna},
    editor = "Liu, Qun  and
      Schlangen, David",
    booktitle = "Proceedings of the 2020 Conference on Empirical Methods in Natural Language Processing: System Demonstrations",
    month = oct,
    year = "2020",
    address = "Online",
    publisher = "Association for Computational Linguistics",
    url = "https://aclanthology.org/2020.emnlp-demos.7/",
    doi = "10.18653/v1/2020.emnlp-demos.7",
    pages = "46--54"
}

@article{wolf2019huggingfacetransformers,
  author       = {Thomas Wolf and
                  Lysandre Debut and
                  Victor Sanh and
                  Julien Chaumond and
                  Clement Delangue and
                  Anthony Moi and
                  Pierric Cistac and
                  Tim Rault and
                  R{\'{e}}mi Louf and
                  Morgan Funtowicz and
                  Jamie Brew},
  title        = {HuggingFace's Transformers: State-of-the-art Natural Language Processing},
  journal      = {CoRR},
  volume       = {abs/1910.03771},
  year         = {2019},
  url          = {http://arxiv.org/abs/1910.03771},
  eprinttype    = {arXiv},
  eprint       = {1910.03771},
  bibsource    = {dblp computer science bibliography, https://dblp.org}
}

@inproceedings{davari2024breadcrumbs,
author = {Davari, MohammadReza and Belilovsky, Eugene},
title = {Model Breadcrumbs: Scaling Multi-task Model Merging with Sparse Masks},
year = {2024},
isbn = {978-3-031-73225-6},
publisher = {Springer-Verlag},
address = {Berlin, Heidelberg},
url = {https://doi.org/10.1007/978-3-031-73226-3_16},
doi = {10.1007/978-3-031-73226-3_16},
booktitle = {Computer Vision – ECCV 2024: 18th European Conference, Milan, Italy, September 29–October 4, 2024, Proceedings, Part LXXV},
pages = {270–287},
numpages = {18},
location = {Milan, Italy}
}

@article{horoi2025less-is-more,
  title={Less is More: Undertraining Experts Improves Model Upcycling},
  author={Stefan Horoi and Guy Wolf and Eugene Belilovsky and Gintare Karolina Dziugaite},
  journal={arXiv preprint arXiv:2506.14126},
  year={2025},
  url={http://arxiv.org/abs/2506.14126}
}

@inproceedings{
zhou2025on-task-vectors-gradients,
title={On Task Vectors and Gradients},
author={Luca Zhou and Daniele Solombrino and Donato Crisostomi and Maria Sofia Bucarelli and Giuseppe Alessio D'Inverno and Fabrizio Silvestri and Emanuele Rodol{\`a}},
booktitle={UniReps: 3rd Edition of the Workshop on Unifying Representations in Neural Models},
year={2025},
url={https://openreview.net/forum?id=747FYd9Oj9}
}

@article{yang2026model_merging,
author = {Yang, Enneng and Shen, Li and Guo, Guibing and Wang, Xingwei and Cao, Xiaochun and Zhang, Jie and Tao, Dacheng},
title = "Model Merging in {LLM}s, {MLLM}s, and Beyond: Methods, Theories, Applications, and Opportunities",
year = {2026},
issue_date = {June 2026},
publisher = {Association for Computing Machinery},
address = {New York, NY, USA},
volume = {58},
number = {8},
issn = {0360-0300},
url = {https://doi.org/10.1145/3787849},
doi = {10.1145/3787849},
journal = {ACM Comput. Surv.},
month = feb,
articleno = {216},
numpages = {41}
}

@article{intellect1,
  author       = {Sami Jaghouar and
                  Jack Min Ong and
                  Manveer Basra and
                  Fares Obeid and
                  Jannik Straube and
                  Michael Keiblinger and
                  Elie Bakouch and
                  Lucas Atkins and
                  Maziyar Panahi and
                  Charles Goddard and
                  Max Ryabinin and
                  Johannes Hagemann},
  title        = {{INTELLECT-1} Technical Report},
  journal      = {CoRR},
  volume       = {abs/2412.01152},
  year         = {2024},
  url          = {https://doi.org/10.48550/arXiv.2412.01152},}

@misc{lidin2026covenant72bpretraining72bllm,
      title={Covenant-72B: Pre-Training a 72B LLM with Trustless Peers Over-the-Internet}, 
      author={Joel Lidin and Amir Sarfi and Erfan Miahi and Quentin Anthony and Shivam Chauhan and Evangelos Pappas and Benjamin Thérien and Eugene Belilovsky and Samuel Dare},
      year={2026},
      eprint={2603.08163},
      archivePrefix={arXiv},
      primaryClass={cs.DC},
      url={https://arxiv.org/abs/2603.08163}, 
}

@misc{ahn2025diondistributedorthonormalizedupdates,
      title={Dion: Distributed Orthonormalized Updates}, 
      author={Kwangjun Ahn and Byron Xu and Natalie Abreu and Ying Fan and Gagik Magakyan and Pratyusha Sharma and Zheng Zhan and John Langford},
      year={2025},
      eprint={2504.05295},
      archivePrefix={arXiv},
      primaryClass={cs.LG},
      url={https://arxiv.org/abs/2504.05295}, 
}

@article{bjork1971iterative_algorithm,
  author  = {Bj{\"o}rck, {\AA}ke and Bowie, Clazett},
  title   = {An Iterative Algorithm for Computing the Best Estimate of an Orthogonal Matrix},
  journal = {SIAM Journal on Numerical Analysis},
  volume  = {8},
  number  = {2},
  pages   = {358--364},
  year    = {1971},
  doi     = {10.1137/0708036},
  url     = {https://doi.org/10.1137/0708036}
}

@article{kovarik1970iterative_methods,
  author  = {Kovarik, Zdislav},
  title   = {Some Iterative Methods for Improving Orthonormality},
  journal = {SIAM Journal on Numerical Analysis},
  volume  = {7},
  number  = {3},
  pages   = {386--389},
  year    = {1970},
  doi     = {10.1137/0707031},
  url     = {https://doi.org/10.1137/0707031}
}

@misc{ai2025practicalefficiencymuonpretraining,
      title={Practical Efficiency of Muon for Pretraining}, 
      author={Essential AI and : and Ishaan Shah and Anthony M. Polloreno and Karl Stratos and Philip Monk and Adarsh Chaluvaraju and Andrew Hojel and Andrew Ma and Anil Thomas and Ashish Tanwer and Darsh J Shah and Khoi Nguyen and Kurt Smith and Michael Callahan and Michael Pust and Mohit Parmar and Peter Rushton and Platon Mazarakis and Ritvik Kapila and Saurabh Srivastava and Somanshu Singla and Tim Romanski and Yash Vanjani and Ashish Vaswani},
      year={2025},
      eprint={2505.02222},
      archivePrefix={arXiv},
      primaryClass={cs.LG},
      url={https://arxiv.org/abs/2505.02222}, 
}
\bibliographystyle{abbrv}


\newpage
\appendix
\crefalias{section}{appendix}
\section{Optimal Hyperparameters}
\label{app:optimal-hparams}

We report the optimal hyperparameters used for each main-text result table: inner learning rate $\eta_{\mathrm{in}}$, outer learning rate $\eta_{\mathrm{out}}$, and outer momentum $\mu$. For the AdamW DP baselines we found optimal learning rates of 0.00195 and 0.00276; weight decays of 0.15 and 0.125 for the 178M and 512M models, respectively.

\begin{table}[h]
\centering
\small
\caption{Optimal hyperparameters for Table~\ref{tab:benchmarking_merging_methods}.}
\label{tab:hparams-merging-methods}
\begin{tabular}{lcc}
\toprule
\textbf{Method} & $\boldsymbol{\eta_{\mathrm{in}}}$ & $\boldsymbol{\eta_{\mathrm{out}}}$ \\
\midrule
DiLoCo SGD & 0.00138 & 1.6 \\
\midrule
TIES    & 0.00276 & 0.7 \\
DARE    & 0.00276 & 1.5 \\
TSV-M   & 0.00276 & 0.6 \\
Iso-C   & 0.00391 & 2.1 \\
Iso-CTS & 0.00391 & 3.1 \\
\bottomrule
\end{tabular}
\end{table}


\begin{table}[h]
\centering
\small
\caption{Optimal hyperparameters for Table~\ref{tab:valid-loss-workers} at 178M scale.}
\label{tab:hparams-workers-178m}
\begin{adjustbox}{width=\linewidth}
\begin{tabular}{llcccccccc}
\toprule
\textbf{Method} & \textbf{Hyperparameter} 
& $\boldsymbol{R=1}$ & $\boldsymbol{R=2}$ & $\boldsymbol{R=4}$ & $\boldsymbol{R=8}$ & $\boldsymbol{R=16}$ 
& $\boldsymbol{R=32}$ & $\boldsymbol{R=64}$ & $\boldsymbol{R=128}$ \\
\midrule
\multirow{3}{*}{DiLoCo SGD}
& $\eta_{\mathrm{in}}$  & 0.00195 & 0.00195 & 0.00276 & 0.00138 & 0.00195 & 0.00138 & 0.00138 & 0.000977 \\
& $\eta_{\mathrm{out}}$ & 1.2 & 1.4     & 1.4     & 1.6     & 1.5     & 1.6     & 1.6     & 1.6 \\
\midrule
\multirow{3}{*}{DiLoCo}
& $\eta_{\mathrm{in}}$  & 0.00138 & 0.00138 & 0.00138 & 0.00138 & 0.00138 & 0.00195 & 0.00276 & 0.00195 \\
& $\eta_{\mathrm{out}}$ & 0.5 & 0.6     & 0.6     & 0.7     & 0.8     & 1       & 0.9     & 1 \\
& $\mu$                & 0.8 & 0.8     & 0.8     & 0.9     & 0.9     & 0.9     & 0.9     & 0.9 \\
\midrule
\multirow{3}{*}{DiLoCo SGD + Iso-C}
& $\eta_{\mathrm{in}}$  & 0.00391 & 0.00391 & 0.00391 & 0.00391 & 0.00138 & 0.000488 & 0.000244 & 0.000122 \\
& $\eta_{\mathrm{out}}$ & 1.7 & 1.7     & 2.1     & 2.1     & 4.6     & 9       & 15.9    & 25.5 \\
\midrule
\multirow{3}{*}{IsoLoCo}
& $\eta_{\mathrm{in}}$  & 0.00195 & 0.00138 & 0.00138 & 0.00138 & 0.00195 & 0.000345 & 0.000244 & 0.000122 \\
& $\eta_{\mathrm{out}}$ & 1.6 & 1.7     & 1.8     & 1.8     & 1.8     & 4.6      & 5.6     & 9.8 \\
& $\mu$                & 0.7 & 0.8     & 0.8     & 0.8     & 0.8     & 0.9      & 0.9     & 0.9 \\
\bottomrule
\end{tabular}
\end{adjustbox}
\end{table}


\begin{table}[h]
\centering
\small
\caption{Optimal hyperparameters for Table~\ref{tab:valid-loss-workers} at 512M scale.}
\label{tab:hparams-workers-512m}
\begin{adjustbox}{width=\linewidth}
\begin{tabular}{llccccccc}
\toprule
\textbf{Method} & \textbf{Hyperparameter}
& $\boldsymbol{R=1}$ & $\boldsymbol{R=2}$ & $\boldsymbol{R=4}$ & $\boldsymbol{R=8}$ 
& $\boldsymbol{R=16}$ & $\boldsymbol{R=32}$ & $\boldsymbol{R=64}$ \\
\midrule
\multirow{3}{*}{DiLoCo}
& $\eta_{\mathrm{in}}$  & 0.00195 & 0.00138 & 0.00195 & 0.000691 & 0.000691 & 0.000488 & 0.000488 \\
& $\eta_{\mathrm{out}}$ & 0.5 & 0.6 & 0.6 & 0.7 & 0.8 & 1 & 0.9 \\
& $\mu$                & 0.8 & 0.8 & 0.8 & 0.9 & 0.9 & 0.9 & 0.9 \\
\midrule
\multirow{3}{*}{IsoLoCo}
& $\eta_{\mathrm{in}}$ & 0.00195 & 0.00138 & 0.00195 & 0.00138 & 0.00195 & 0.000345 & 0.000345 \\
& $\eta_{\mathrm{out}}$ & 1.6 & 1.7 & 1.8 & 1.8 & 1.8 & 4.6 & 5.6 \\
& $\mu$                & 0.7 & 0.8 & 0.8 & 0.8 & 0.8 & 0.9 & 0.9 \\
\bottomrule
\end{tabular}
\end{adjustbox}
\end{table}


\begin{table}[h!]
\centering
\small
\caption{Optimal hyperparameters for Table~\ref{tab:inner-steps-scaling}.}
\label{tab:hparams-inner-steps}
\begin{tabular}{llcccc}
\toprule
\textbf{Method} & \textbf{Hyperparameter}
& $\boldsymbol{H=30}$ & $\boldsymbol{H=60}$ & $\boldsymbol{H=120}$ & $\boldsymbol{H=240}$ \\
\midrule
\multirow{3}{*}{DiLoCo}
& $\eta_{\mathrm{in}}$  & 0.00138 & 0.00138 & 0.00195 & 0.00195 \\
& $\eta_{\mathrm{out}}$ & 0.7     & 0.9     & 1       & 1.1 \\
& $\mu$                & 0.9     & 0.8     & 0.8     & 0.7 \\
\midrule
\multirow{3}{*}{IsoLoCo}
& $\eta_{\mathrm{in}}$  & 0.00138 & 0.00138 & 0.000977 & 0.00138 \\
& $\eta_{\mathrm{out}}$ & 1.8      & 1.7      & 1.7      & 1.7 \\
& $\mu$                & 0.8      & 0.8      & 0.8      & 0.7 \\
\bottomrule
\end{tabular}
\end{table}

\newpage
\section{Comparing IsoLoCo to Muon}\label{a:iso_vs_muon}
IsoLoCo bears some similarities with the Muon optimizer, which has been recently proposed as an effective optimizer for 2D parameters in LLMs~\cite{jordan2024muon, liu2025muonscalablellmtraining}. In this section, we discuss the similarities and differences between the two.

\begin{wrapfigure}{r}{0.52\textwidth}
    \vspace{-1em}
    \centering
    \begin{minipage}{0.50\textwidth}
        \hrule height 0.8pt
        \vspace{0.5ex}
        \captionof{algorithm}{Muon}
        \label{alg:muon}
\vspace{-5pt}
        \hrule height 0.4pt
        \vspace{0.75ex}
        \begin{algorithmic}[1]
            \Require Learning rate $\eta$, momentum $\mu$
            \State Initialize momentum buffer $B_0 \gets 0$
            \For{$t = 1, \ldots, T$}
                \State Compute gradient $G_t \gets \nabla_{\theta}\mathcal{L}_t(\theta_{t-1})$
                \State $B_t \gets \mu B_{t-1} + G_t$
                \State $O_t \gets \operatorname{NewtonSchulz}(B_t)$
                \State $O_t \gets \sqrt{\text{fan\_out}/\text{fan\_in}} \cdot O_t$
                \State Update parameters $\theta_t \gets \theta_{t-1} - \eta O_t$
            \EndFor
            \State \Return $\theta_T$
        \end{algorithmic}
        \vspace{0.75ex}
        \hrule height 0.8pt
    \end{minipage}
    \vspace{-1em}
\end{wrapfigure}

Muon approximately orthogonalizes the parameter gradients through the use of the Newton-Schulz iterative method described in \Cref{ss:fast_isoloco}. Muon then scale the gradients by $\sqrt{\text{fan\_out}/\text{fan\_in}}$ so that the effective magnitude of the gradients is consistent across layers of different sizes~\cite{yang2024spectralconditionfeaturelearning}. We will call this normalization here ``spectral''.

The main difference between IsoLoCo is that IsoLoCo actually computes the SVD of the gradient matrices, giving us access to the singular values and allowing to set them all to their mean. On the other hand, Muon performs Newton-Shulz iterations to approximately set all singular values to 1 and then normalizes then by $\sqrt{\text{fan\_out}/\text{fan\_in}}$.
Since we also use Newton--Schulz in the fast implementation of IsoLoCo, the main difference becomes the scaling: IsoLoCo scales the orthogonalized gradient matrices by the mean/$\operatorname{RMS}$ of their singular values while Muon scales by $\sqrt{\text{fan\_out}/\text{fan\_in}}$. 

\begin{wraptable}{r}{0.5\textwidth}
\vspace{-1em}
\centering
\small
\caption{Validation loss for scaling choices under different orthogonalization methods.}
\label{tab:iso_muon_ablation}
\begin{tabular}{lcc}
\toprule
& \multicolumn{2}{c}{\textbf{Orthog. Method}} \\
\cmidrule(lr){2-3}
& \textbf{SVD} & \textbf{Newton-Schulz} \\
\midrule
Mean/RMS scale & 2.898 & 2.902 \\
$\sqrt{\frac{\texttt{fan\_out}}{\texttt{fan\_in}}}$ scale & -- & 3.005 \\
\bottomrule
\end{tabular}
\vspace{-1em}
\end{wraptable}

We isolate the impact of the orthogonalization (SVD vs. Newton--Shulz) and the scaling (mean vs spectral) on IsoLoCo by implementing these variants and keeping all other factors constant. We report results for our 178M models with $R=8$ and $H=30$ in \Cref{tab:iso_muon_ablation}. The results with SVD orthogonalization and mean scaling (IsoLoCo) and with Newton--Schulz orthogonalization and RMS scaling (fast implementation of IsoLoCo) were already presented in the main text, with both variants performing well. On the other hand, Newton--Schulz orthogonalization with $\sqrt{\text{fan\_out}/\text{fan\_in}}$ scaling, which roughly represents Muon as a DiLoCo outer optimizer, is significantly worse than IsoLoCo, with 3.4\% higher loss.

Other notable differences between IsoLoCo and Muon are the fact that Muon uses standard momentum while IsoLoCo uses Nesterov. Also, Muon applies the Newton--Schulz-based orthogonalization to the final update matriecs $B_t$ while IsoLoCo applies the orthogonalization to the gradients directly. Both of these variations were found to perform worse than our IsoLoCo design (see \Cref{ss:ablations}).
Another difference is in how these two methods treat 1D parameters, IsoLoCo simply averages them and applies momentum (similar to SGD + Nesterov optimizer) while Muon typically optimizes those with another optimizer such as AdamW. Lastly, it is common in Muon implementations to exclude the 2D embedding and head parameters from the orthogonalization step, we did not find this to significantly affect performance.

\end{document}